\pdfoutput=1

\documentclass[11pt]{article}

\usepackage[]{acl}

\usepackage{times}
\usepackage{latexsym}

\usepackage[T1]{fontenc}

\usepackage[utf8]{inputenc}

\usepackage{microtype}
\usepackage{multirow}
\usepackage{makecell}
\usepackage{graphicx}

\usepackage{booktabs}
\usepackage{amssymb}

\DeclareMathSymbol{\mh}{\mathord}{operators}{`\-}

\title{LongT5: Efficient Text-To-Text Transformer for Long Sequences}

\author{Mandy Guo\thanks{\hspace{2mm}Equal contributions.} \thanks{\hspace{2mm} Corresponding authors.},~
Joshua Ainslie\footnotemark[1]    \footnotemark[2],~
David Uthus\footnotemark[1],~ Santiago Onta\~{n}\'{o}n\footnotemark[1]\\
{\bf Jianmo Ni,}~
{\bf Yun{-}Hsuan Sung,}~
{\bf Yinfei Yang}
  \AND
  {\rm Google Research}\\
 \{xyguo, jainslie, duthus, santiontanon, jianmon, yhsung, yinfeiy\}@google.com
}

\begin{document}
\maketitle
\begin{abstract}
Recent work has shown that either (1) increasing the input length or (2) increasing model size can improve the performance of Transformer-based neural models. In this paper, we present {\em LongT5}, a new model that explores the effects of scaling both the input length and model size at the same time. 
Specifically, we integrate attention ideas from long-input transformers (ETC), and adopt pre-training strategies from summarization pre-training (PEGASUS) into the scalable T5 architecture. The result is a new attention mechanism we call {\em Transient Global} (TGlobal), which mimics ETC's local/global attention mechanism, but without requiring additional side-inputs. We are able to achieve state-of-the-art results on several summarization and question answering tasks, as well as outperform the original T5 models on these tasks. We have open sourced our architecture and training code, as well as our pre-trained model checkpoints.
\end{abstract}

\section{Introduction}

Transformer models such as BERT~\cite{devlin-etal-2019-bert}, and other variants~\cite{liu2019roberta,Radford2019LanguageMA,t5,lewis-etal-2020-bart} have achieved state-of-the-art results on many challenging NLP tasks. 
Moreover, recent work in long-input transformers~\cite{etc,zaheer2020,Beltagy2020Longformer,tay2021long} has shown that increasing the input length a Transformer is able to process results in further performance gains. Additionally, it is also known that increasing model size also leads to performance gains in many tasks~\cite{kaplan2020scaling}. 

\begin{figure}[t]
  \centering
  \includegraphics[width=\linewidth]{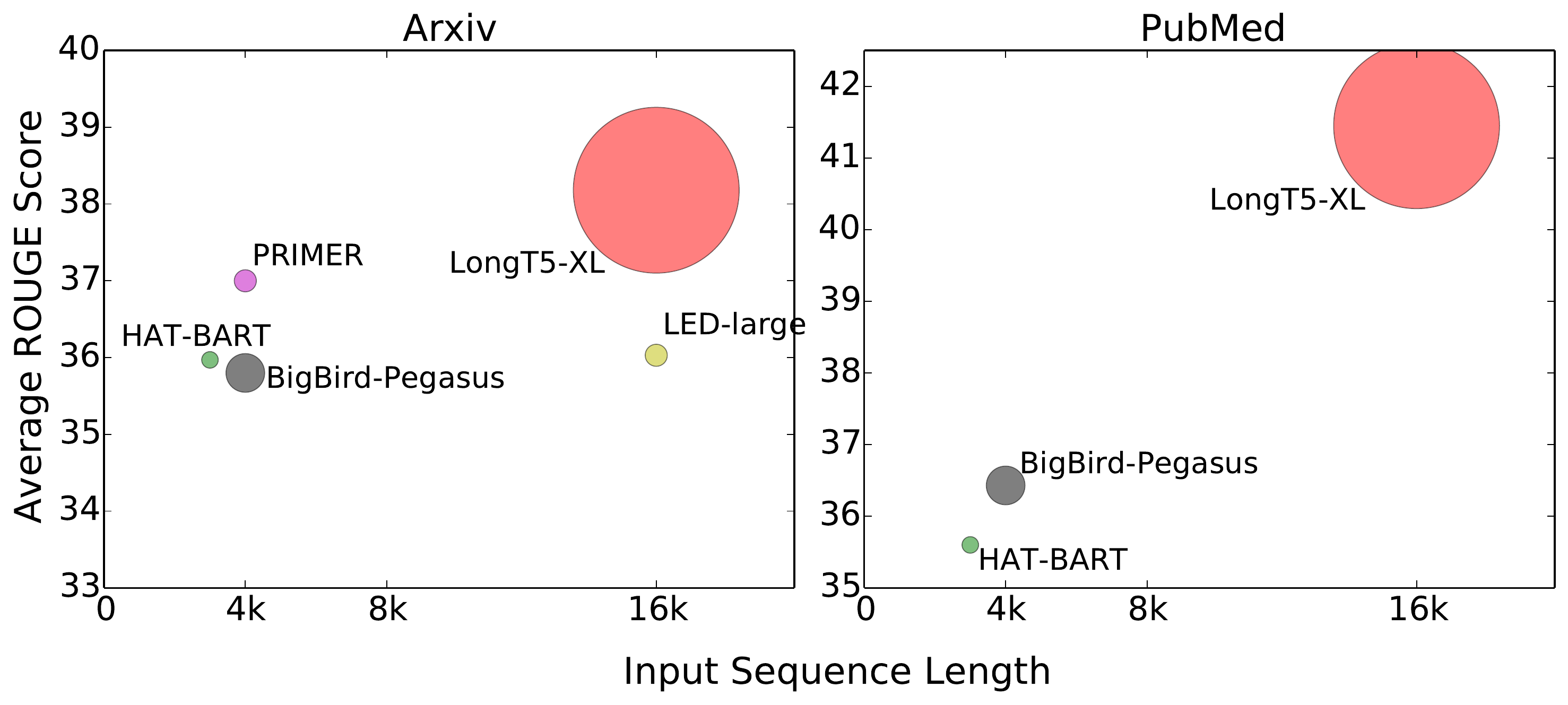}
  \caption{The average ROUGE score~($(R\mh1+R\mh2+R\mh{L})/3$) of LongT5 and baseline models on arXiv and PubMed summarization tasks~\cite{cohan-etal-2018-discourse} with different input length~($x$ axis). Baseline models: HAT-BART~\cite{rohde2021hierarchical}, BigBird-PEGASUS~\cite{zaheer2020}, PRIMER~\cite{xiao2021primer}, LED~\cite{Beltagy2020Longformer}. The size of circle roughly indicates the $\#$ of parameters for each model.
  }
  \label{fig:gtr_avg}
\end{figure}

In this paper, we present a new model, called {\em LongT5}, with which we explore the effects of scaling both the input length and model size at the same time.
To achieve this, we integrate long-input transformer attention and pre-training ideas into the scalable T5~\cite{t5} model architecture.
The resulting model, as shown in Figure~\ref{fig:gtr_avg}, achieves state-of-the-art performance on several tasks which require handling long sequence inputs.

Regarding attention, we design a new attention mechanism, which we call {\em Transient Global} (TGlobal), that mimics ETC's local/global mechanism~\cite{etc}. Importantly, TGlobal attention removes the need for the additional side inputs in ETC, in order to fit within the T5 architecture. The main idea of ETC's local/global mechanism is to introduce local sparsity in the attention mechanism to reduce the quadratic cost when scaling to long inputs. Specifically, ETC only allows tokens in the input (called the {\em long input}) to attend to a local neighborhood, and adds a secondary input called the {\em global memory}, through which tokens in the long input can attend to each other indirectly. One disadvantage of this mechanism is that it requires designing this secondary global input for each new problem. In order to adapt it to T5, our new TGlobal mechanism {\em synthesizes} these global tokens on the fly (as aggregations of groups of tokens in the input), at each attention layer. Our experiments show that this mechanism results in only a small degradation in performance with respect to full attention in the same input length but allows the model to scale to much larger input lengths, resulting in significant performance gains.

Regarding pre-training, we adopt the pre-training strategy in the PEGASUS~\cite{pegasus} model. This pre-training strategy was originally designed for abstractive summarization, but in our experiments, we found it also improves model performance for other tasks, such as question answering, and hence we adopted it in LongT5. The key idea is to mask out key (principle) sentences from a document and ask the model to reproduce them as a single string, as if it was a summary.


We evaluate LongT5 on several summarization and question answering tasks (see Sections \ref{subsec:summarization-datasets} and \ref{subsec:qa-datasets} for detailed descriptions of these datasets). Thanks to the scaling of both input length and model size, we achieve state-of-the-art results on many of them.

The main contributions of this work are:
\begin{itemize}
    \item A new Transformer architecture, {\em LongT5}, that allows for scaling {\em both} input length and model scale at the same time.
    \item A new attention mechanism (TGlobal), which mimics ETC's local/global mechanism but is a drop-in replacement to regular attention for existing Transformer architectures like T5.
    \item An analysis of model performance when varying both input length and model size of vanilla T5 and LongT5 models (pushing both models up to the maximum lengths they can handle before encountering memory issues), to understand the trade-offs in both performance and computation cost.
    \item State-of-the-art results on the arXiv, PubMed, BigPatent, MediaSum, and TriviaQA datasets. For Natural Questions, we used a slightly different formulation than the original tasks, and hence we do not make state-of-the-art claims.
    \item We open source our model architecture\footnote{Published under the Flaxformer GitHub \url{https://github.com/google/flaxformer/tree/main/flaxformer/architectures/longt5}} and training code, as well as pre-trained model checkpoints on GitHub\footnote{\url{https://github.com/google-research/longt5}}.
\end{itemize}

\section{T5}\label{sec:t5}
T5 \cite{t5} is a transformer based text-to-text pre-trained language model that is gaining popularity for its unified framework that converts all text-based language problems into a text-to-text format, and its ease to scale up in number of parameters (from 60M to 11B parameters) with model parallelism. With full attention transformer, T5 has been successfully applied to many NLP tasks, but the tasks only require shorter input sequences. This is due to the limitation of quadratic computation growth with respect to input sequence length, resulting in larger memory consumption and longer training time. Recently, \citet{trainshort_testlong} explored scaling up T5 style models at inference time to longer sequences than seen during training, but how to scale up T5 style models in the input sequence length during training remains underexplored.

\begin{figure*}[t!]
	\includegraphics[width=0.84\textwidth]{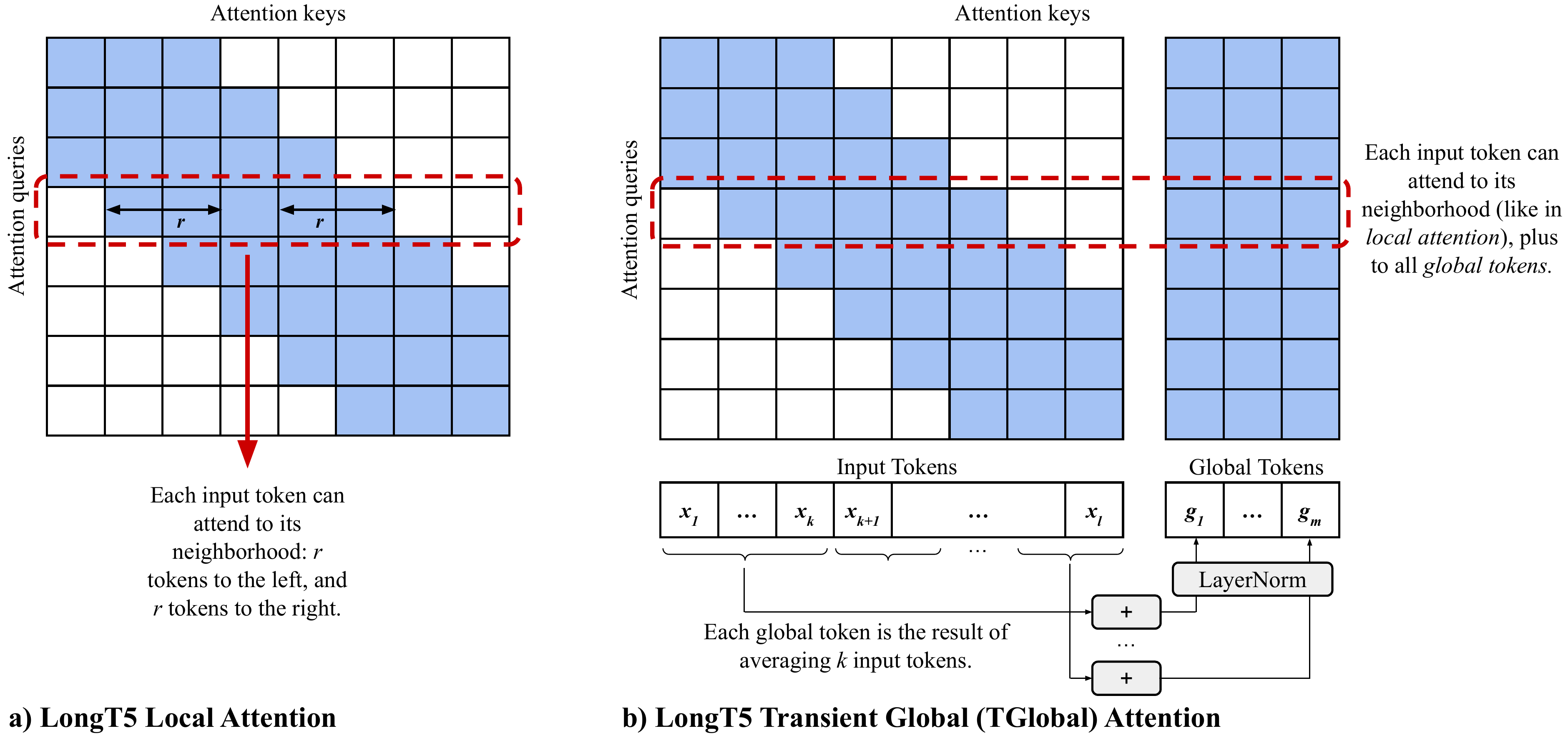}
	\centering
	\caption{Illustration of the two attention mechanisms we experimented with in LongT5.}
	\label{fig:attention}
\end{figure*}

\section{LongT5}\label{sec:longt5}
\subsection{Architecture}
We extend the original T5 encoder with global-local attention sparsity patterns~\cite{etc,bigbird} to handle long inputs. For the work reported in this paper, we used a standard T5 decoder since all of the tasks we considered require relatively short output sequence lengths.

Architecturally, the main difference between T5 and LongT5 lies in the attention mechanism. We experiment with two attention mechanism variations for LongT5, illustrated in Figure \ref{fig:attention}: (1) \emph{Local Attention} and (2) \emph{Transient Global Attention (TGlobal)}.  Both variations preserve several properties of T5: relative position representations, support for example packing, and compatibility with T5 checkpoints.

\subsubsection{Local Attention}
For {\em Local Attention}, we simply replace the encoder self-attention operation in T5 with a sparse sliding-window local attention operation following the implementation in ETC \cite{etc}.  Specifically, for a given local radius $r$, this formulation only allows each token to attend $r$ tokens to the left and right of it (see Figure \ref{fig:attention}.a).  We found $r = 127$ to be sufficient in practice, where $r$ is the number of neighboring tokens to the left and to the right. 

Local Attention does not introduce any new parameters and easily accommodates the attention masking required for example packing\footnote{Example packing refers to packing more than one short example in the same input sequence to increase training efficiency. This is specially useful in LongT5, since with the large input lengths used in our model, if many examples are short, most of the input sequence would be dedicated to padding, wasting significant computation.}. For a given choice of $r$, complexity is linear in input sequence length $l$: $O(l \times r)$.

\subsubsection{Transient Global Attention (TGlobal)}
To allow input tokens to interact with each other in each layer of the encoder at a longer range than Local Attention’s local radius, we introduce {\em Transient Global Attention} as a modification of ETC’s global-local attention in a ``fixed blocks’’ pattern.  Namely, we divide the input sequence into blocks of $k$ tokens, and for each block we compute a global token by summing (and then normalizing) the embeddings of every token in the block (see Figure \ref{fig:attention}.b).  Now when computing attention, we allow each input token to attend not only to nearby tokens like in Local Attention, but also to every global token. We call these global tokens \emph{transient} because in contrast to ETC-like global-local attention patterns, these tokens are dynamically constructed (and subsequently discarded) within each attention operation, removing any requirement for deciding which input tokens should be treated as ``global’’.

TGlobal attention only introduces a couple new parameters\footnote{For base models, we introduced 10k additional parameters, 25k for large, and 50k for xl.}: (1) T5-style relative position biases representing the distance from an input token’s block to the block of each global token it’s attending to, and (2) T5-style layer normalization parameters for normalizing each global token’s embedding.  The rest of the parameters are identical to T5, and we accommodate sequence packing by additionally masking attention from input tokens to global tokens of other examples.  We found block size $k = 16$ to be sufficient in practice. Notice thus, that TGlobal attention introduces a block of $l * l/k$ additional attention key-value pairs to calculate on top of Local Attention ($l$ input tokens, attending to $l/k$ global tokens; represented by the right most rectangle in Figure \ref{fig:attention}.b), hence
for input sequence length $l$, complexity is $O(l(r + l/k))$.



\subsection{PEGASUS Principle Sentences Generation Pre-training}
\label{sec:pegasus}
T5 is pre-trained with a span corruption objective, where spans of consecutive input tokens are replaced with a mask token and the model is trained to reconstruct the masked-out tokens.
While it is effective, recent work on masked language modeling (MLM)~\cite{liu2019roberta,zhang-etal-2019-ernie} shows that carefully selecting the prediction objective could lead to significantly better performance. 
One argument is that predicting more informative tokens from the text could force the model to learn better semantics of the text. 
Motivated by that, we explore masking and generating the principle sentences from the text.
In particular, we adopt the Gap Sentences Generation with \textbf{Principle Ind-Uniq} strategy from \citet{pegasus}, which was used for summarization pre-training.

Following \citet{pegasus}, we select top-$m$ scored~(\textbf{Principle}) sentences based on ROUGE-F1 score~\cite{lin-2004-rouge} using $s_i = rouge(x_i, D~\backslash~\{x_i\}, \forall_i)$,
where $i$ is the sentence index, $D$ is the collection of sentences in the document.
Each sentence is scored independently~(\textbf{Ind}), and each $n$-gram is only counted once~(\textbf{Uniq}).


\section{Experiments}\label{sec:experiments}
\subsection{Configurations}
LongT5 is implemented using JAX\footnote{https://github.com/google/jax} and the Flaxformer\footnote{https://github.com/google/flaxformer} library. Following the same setup as T5.1.1\footnote{https://github.com/google-research/text-to-text-transfer-transformer/blob/main/released\_checkpoints.md\#t511}, we consider models of 3 sizes: base ($\sim$220M), large ($\sim$770M), and xl ($\sim$3B), 
and use the same cased English SentencePiece vocab model used by T5.1.1, which contains 32000 sentence pieces. We use batch size of 128 and Adafactor as the optimizer in all experiments. We decide to use greedy decoding instead of beam search for all our experiments even with the test sets, therefore, our results reported below could potentially be improved further by using beam search, but we would like to make the setup consistent with our dev setup.

\subsubsection{Pre-training}
We pre-train LongT5 models for 1M steps on 4096 input sequence length and 910 output sequence length. We use the same inverse square-root learning rate schedule as T5, with learning rate set to $1/\sqrt{max(step, warm\_up\ steps)}$, where warm\_up steps is set to 10000. The same as T5.1.1, we pre-train LongT5 only on the C4 dataset \cite{c4}, and we do not apply dropout during pre-training. As described in section \ref{sec:pegasus}, we use the PEGASUS Principle Sentences Generation objective as our pre-training objective. The configuration is similar to what was described by \newcite{pegasus} for their larger models, except for the masked sentence ratio in which we use a value of 0.2 instead of 0.45\footnote{We briefly experimented with other values, but found 0.2 to work best with the downstream tasks of interest.}. In section \ref{subsec:analysis:pegasus_vs_span}, we will show our ablation study between Principle Sentences Generation and Span Corruption.

\subsubsection{Fine-tuning}
For fine-tuning, we use a constant learning rate of 0.001 and dropout rate of 0.1 for all tasks. 
For summarization tasks, we experiment with values of 4096, 8192, and 16384 for input lengths and 512 for output lengths.
For QA tasks, we experiment with values starting at 512 and scale up to 36864 for input lengths and 128 for output lengths.

\begin{table*}
    \small
    \centering
    \begin{tabular}{ p{2.45cm} | r r r | r r r r } \toprule
    \multirow{2}{*}{\textbf{Dataset}} & \multicolumn{3}{c}{\textbf{Example Count}} & \multicolumn{4}{c}{\textbf{Input Length}}  \\ 
    & Train & Validation & Test & Average & Median & Max & 90th percentile \\ \midrule
    CNN / Daily Mail      & 287,113 & 13,368 & 11,490  & 982.39 & 894 & 5268 & 1659 \\
    arXiv      & 203,037 & 6,436 & 6,440 & 10,720.18 & 8,519 & 378,825 & 20,170 \\
    PubMed  & 119,924 & 6,633 & 6,658 & 4,747.97 & 3,883 & 452,915 & 8,883 \\
    BigPatent   & 1,207,222 & 67,068 & 67,072 & 6,537.32 & 5,236 & 294,004 & 11,328\\
    MediaSum    & 443,596  & 10,000 & 10,000 & 2,302.02 & 1,748 & 125,974 & 4,128 \\
    Multi-News   & 44,972 & 5,622 & 5,622 & 2,593.81 & 1,902.5 & 683,544 & 4,853\\
    \bottomrule
    \end{tabular}
    \caption{Statistics for the summarization datasets. Input length measured in tokens using a SentencePiece Model.}
    \label{tab:summarization_stats}
\end{table*}

\begin{table}
\small
\centering
\begin{tabular}{lccc}
\toprule
& \multicolumn{3}{c}{\textbf{arXiv}} \\
\textbf{Approach} & R-1 & R-2 & R-L \\
\midrule
DANCER PEGASUS & 45.01 & 17.6 & 40.56  \\
BigBird-PEGASUS (large) & 46.63 & 19.02 & 41.77  \\
HAT-BART & 46.68 & 19.07 & 42.17  \\
LED (large) & 46.63 & 19.62 & 41.83   \\ 
PRIMER & 47.6 & 20.8 & 42.6  \\
\midrule
LongT5 (large - 16k input) & 48.28 & 21.63 & 44.11  \\
LongT5 (xl - 16k input) & \textbf{48.35} & \textbf{21.92} & \textbf{44.27} \\
\toprule
&  \multicolumn{3}{c}{\textbf{PubMed}} \\
\textbf{Approach} & R-1 & R-2 & R-L \\
\midrule
DANCER PEGASUS  & 46.34 & 19.97 & 42.42 \\
BigBird-PEGASUS (large)  & 46.32 & 20.65 & 42.33  \\
HAT-BART  & 48.36 & 21.43 & 37.00 \\
\midrule
LongT5 (large - 16k input) & 49.98 & 24.69 & 46.46 \\
LongT5 (xl - 16k input) &  \textbf{50.23} & \textbf{24.76} & \textbf{46.67} \\
\toprule
& \multicolumn{3}{c}{\textbf{BigPatent}} \\
\textbf{Approach} & R-1 & R-2 & R-L\\
\midrule
BigBird-PEGASUS (large) & 60.64 & 42.46 & 50.01 \\
\midrule
LongT5 (large - 16k input) & 70.38 & 56.81 & 62.73  \\
LongT5 (xl - 16k input) & \textbf{76.87} & \textbf{66.06} & \textbf{70.76}  \\
\toprule
& \multicolumn{3}{c}{\textbf{MultiNews}} \\
\textbf{Approach} & R-1 & R-2 & R-L\\
\midrule
TG-MultiSum  & 47.10 & 17.55 & 20.73 \\
PRIMER  & \textbf{49.9} & \textbf{21.1} & \textbf{25.9} \\
\midrule
LongT5 (large - 8k input) & 47.18 & 18.44 & 24.18 \\
LongT5 (xl - 8k input) & 48.17 & 19.43 & 24.94 \\
\toprule
& \multicolumn{3}{c}{\textbf{MediaSum}}  \\
\textbf{Approach} & R-1 & R-2 & R-L \\
\midrule
BART (large) & 35.09 & 18.05 & 31.44  \\
\midrule
LongT5 (large - 4k input) & 35.54 & 19.04 & 32.20 \\
LongT5 (xl - 4k input) & \textbf{36.15} & \textbf{19.66} & \textbf{32.80}\\
\toprule
&  \multicolumn{3}{c}{\textbf{CNN / Daily Mail}} \\
\textbf{Approach} & R-1 & R-2 & R-L\\
\midrule
HAT-BART & \textbf{44.48} & 21.31 & \textbf{41.52} \\
\midrule
LongT5 (large - 4k input) &  42.49 & 20.51 & 40.18 \\
LongT5 (xl - 4k input) &  43.94 & \textbf{21.40} & 41.28 \\
\bottomrule
\end{tabular}
\caption{Summarization results comparing LongT5 with best known approaches. LongT5 scores are with models using TGlobal attention. For each task, we scale up the input length depending on the inputs' statistics, thus not all are scaled to 16k. For more results, please see Section \ref{sec:sum_appendix} in the Appendix.}
\label{tab:summarization_results}
\end{table}

\begin{table*}
    \small
    \centering
    \begin{tabular}{ p{1.45cm} | r r r | r r r r } \toprule
    \multirow{2}{*}{\textbf{Dataset}} & \multicolumn{3}{c}{\textbf{Example Count}} & \multicolumn{4}{c}{\textbf{Input Length}}  \\ 
    & Train & Validation & Test & Average & Median & Max & 90th percentile \\ \midrule
    NQ & 307,373 & 7,830 & & 6,695.92 & 4,486 & 151,519 & 15,290.8 \\
    TriviaQA & 87,622 & 11,313 & 10,832 & 69,082.51 & 45,011 & 1,174,918 & 150,643 \\
    \bottomrule
    \end{tabular}
    \caption{Statistics for the QA datasets. Input length measured in tokens using a SentencePiece Model.}
    \label{tab:qa_stats}
\end{table*}

\begin{table}
\small
\centering
\begin{tabular}{lcc}
\toprule
& \multicolumn{2}{c}{\textbf{NQ}} \\
\textbf{Approach} & EM & F1\\
\midrule
T5.1.1 (base - 512 input) & 50.93 & 52.54 \\
T5.1.1 (base - 6k input) & 56.73 & 56.73 \\
T5.1.1 (large - 512 input) & 57.29 & 60.68  \\
T5.1.1 (large - 3k input) & 60.09 & 64.17 \\
T5.1.1 (xl - 4k input) & 60.75 & 64.07  \\
\midrule
{\bf Local:} \\
LongT5 (base - 512 input) & 54.39 & 58.24\\
LongT5 (base - 36k input) & 55.77 & 59.66 \\
LongT5 (large - 512 input) & 55.19 & 58.00 \\
LongT5 (large - 10k input) & 60.01 & 64.40 \\
{\bf TGlobal:} \\
LongT5 (base - 512 input) & 55.73 & 59.06 \\
LongT5 (base - 12k input) & 58.12 & 62.44 \\
LongT5 (large - 512 input) & 57.55 & 61.53\\
LongT5 (large - 4k input) & 60.77 & 65.38 \\
LongT5 (large - 6k input) & 59.17 & 63.38 \\
LongT5 (xl - 8k input) & {\bf 62.66} & {\bf 66.61} \\
\\
\toprule
& \multicolumn{2}{c}{\textbf{TriviaQA}} \\
\textbf{Approach} & EM & F1\\
\midrule
BigBird-ETC (random attn) & 80.86 & 84.5 \\
Fusion-in-Decoder & 80.09 & 84.35\\
ReadTwice & 76.86 & 80.85 \\
\midrule
{\bf TGlobal:} \\
LongT5 (base - 16k input) & 74.67 & 78.9 \\
LongT5 (large - 16k input) & 78.38 & 82.45 \\
LongT5 (xl - 16k input) & {\bf 81.00} & {\bf 84.83}  \\
\bottomrule
\end{tabular}
\caption{QA results: (1) NQ results comparing T5.1.1 and LongT5. Base/large models are trained on 4x8 TPUv3 with no model partitioning. Xl models are trained on 8x16 TPUv3 with 8 partitions. (2) TriviaQA results compared to top models on leader board. LongT5 scores using Local and TGlobal attention. Full results in Appendix \ref{sec:qa_appendix}.}
\label{tab:QA_results}
\end{table} 

\subsection{Evaluation on Summarization Tasks}
We choose to benchmark our models on summarization tasks that cover various context lengths, because of their long context understanding and generative nature.

\subsubsection{Datasets}\label{subsec:summarization-datasets}
LongT5 was benchmarked on the following six datasets. 
\vspace{-1mm}
\paragraph{CNN / Daily Mail} \cite{nallapati-etal-2016-abstractive} News from CNN and Daily Mail are used as input and the article's summary bullets are the target summary.
\vspace{-1mm}
\paragraph{PubMed}\cite{cohan-etal-2018-discourse} Scientific documents were collected from PubMed, with a document's content used as input and its corresponding abstract as the target summary.
\vspace{-1mm}
\paragraph{arXiv} \cite{cohan-etal-2018-discourse} Similar to PubMed, but with documents taken from arXiv.
\vspace{-1mm}
\paragraph{BigPatent} \cite{sharma-etal-2019-bigpatent} U.S. patent documents, with the patent's details used as input and the patent's abstract as the target summary.
\vspace{-1mm}
\paragraph{MediaSum} \cite{zhu-etal-2021-mediasum} Interview transcripts from CNN and NPR were used as input and their corresponding topic and overviews used as the target summary.
\vspace{-1mm}
\paragraph{Multi-News} \cite{fabbri-etal-2019-multi} The task involves summarizing multiple news documents about a topic into a human-written summary.

Table \ref{tab:summarization_stats} provides statistics for the number of examples in train, validation, and test splits, and the average, median, max, and 90th percentile input sequence length.
As can be seen, these datasets are long in input length, and would benefit from models that can model lengthier inputs.
We included the CNN / Daily Mail dataset to benchmark on a common task, especially to see how using TGlobal attention impacts the model, despite the length of the inputs being smaller than the other datasets.

\subsubsection{Results}
 
We compare LongT5 with various top approaches: BigBird-PEGASUS \cite{zaheer2020}, HAT-BART \cite{rohde2021hierarchical}, DANCER PEGASUS \cite{gidiotis2020}, PRIMER \cite{xiao2021primer}, TG-MultiSum \cite{cui2021topicguided}, LED \cite{Beltagy2020Longformer}, and an application of BART by \newcite{zhu-etal-2021-mediasum}. For these comparisons, we use common evaluation metrics of ROUGE-1, ROUGE-2, and ROUGE-L.

As can be seen in Table \ref{tab:summarization_results}, LongT5 is able to achieve state-of-the-art rouge scores for arXiv, PubMed, BigPatent, and MediaSum.
For arXiv and PubMed, which are composed of longer inputs, being able to scale up to 16k input length helps LongT5 achieve strong results.

One dataset where LongT5 is not able to achieve state-of-the-art results is with Multi-News.
LongT5 is the 2nd best model, slightly worth than PRIMER.
This is understandable as the PRIMER model was pre-trained on a large corpus of documents related to news events, thus exposing the model to a similar corpus as that seen in Multi-News.

When looking at CNN / Daily Mail, we can see that LongT5 was comparable with HAT-BART, despite not having full attention.
LongT5 did at least get stronger scores in the ROUGE-2 metric.

\subsection{Evaluation on QA Tasks}
For the evaluation on QA tasks, we choose two popular benchmarks, Natural Questions and TriviaQA, that require long context understanding.  

\subsubsection{Datasets}\label{subsec:qa-datasets}
\vspace{-1mm}
\paragraph{NaturalQuestions (NQ)} Questions are real queries issued by multiple users to Google search that retrieve a Wikipedia page in the top five search results. Answer text is drawn from the search results \cite{naturalquestions}.

The original NQ dataset asks models to predict a short answer (including no-answer or yes/no) and a long answer. We framed the task as a seq2seq task and ignored the long answer. Hence, our results focus only on short answer. Moreover, since our models predict answer texts instead of answer spans, our evaluation method differs slightly from the leader boards, and our results are not directly comparable to other existing approaches: (1) Since only the train and dev sets are publicly available, we use $90\%$ of the official train set for training while using $10\%$ as hold-out dev set to fine-tune the hyperparameters and training epoch, and use the official dev set as our test set. (2) We benchmark LongT5 against the corresponding T5.1.1 models instead of directly comparing to the leader boards.
\vspace{-1mm}
\paragraph{TriviaQA} Trivia enthusiasts authored question-answer pairs.
Answers are drawn from Wikipedia and Bing web search results, excluding trivia websites \cite{triviaqa}.

We use the official train/validation splits for training and fine-tuning the hyperparameters and training epoch, then re-train that model combining both train and validation sets to evaluate on the Wikipedia domain on the leader board \footnote{https://competitions.codalab.org/competitions/17208}.


Table \ref{tab:qa_stats} shows the dataset statistics for the number of examples in train and validation splits, and the average, median, max, and 90th percentile input sequence length.
\subsubsection{Results}

Table \ref{tab:QA_results} shows a summary of the results for the NQ and TriviaQA datasets (see Appendix \ref{sec:qa_appendix} for full results). For each dataset, we show two metrics: EM (Exact Match) and F1 score (evaluating precision and recall of individual words in the answer compared to the ground truth, ignoring stop words). 

For NQ, we compare T5.1.1, LongT5 with Local Attention, and LongT5 with TGlobal attention. We decided to run T5.1.1 (1) with the default 512 input sequence length\footnote{For base and large models.} and (2) with the largest input sequence length that can fit into device memory\footnote{For base and large models, we used 4x8 TPUv3 and no model partitioning; for xl model, we used 8x16 TPUv3 and 8 partitions.}, and use those as baselines. Since we are comparing against T5.1.1, for LongT5 experiments we report results at 512 input length for base and large, and the largest input length allowed by each model before running out of memory on the same hardware configuration used in our T5.1.1 experiments.

As the table shows, increasing input length generally results in significant benefits in NQ, with models with larger input lengths significantly outperforming those with smaller input lengths in most cases. Some times, models with the largest input lengths underperform those with 4k length, but we believe those to be due to noise in the experiments, as results are the output of just one repetition of each experiment due to resource constraints. Moreover,
while LongT5 with Local Attention often underperforms T5.1.1, LongT5 with TGlobal attention significantly outperforms T5.1.1. For example, considering the {\em large} size models, T5.1.1 was able only to scale up to an input length of 3k tokens, while the TGlobal model was able to reach 6k tokens, outperforming T5.1.1 at 4k token length (there was a dip at 6k token length, but we hypothesize this is just due to variance, as we only did one run for each configuration).

For TriviaQA, we compare LongT5 with various top approaches on the leader board: BigBird-ETC \cite{bigbird}, Fusion-in-Decoder \cite{fid}, and ReadTwice \cite{readtwice}. As shown in Table \ref{tab:qa_stats}, TriviaQA inputs are quite long, therefore being able to scale up both in model size and to 16k input length helps LongT5 achieve state-of-the-art.

\section{Analysis}\label{sec:analysis}


\subsection{Input Length vs Speed}\label{subsec:analysis:speed}

\begin{figure}[t!]
	\includegraphics[width=\columnwidth]{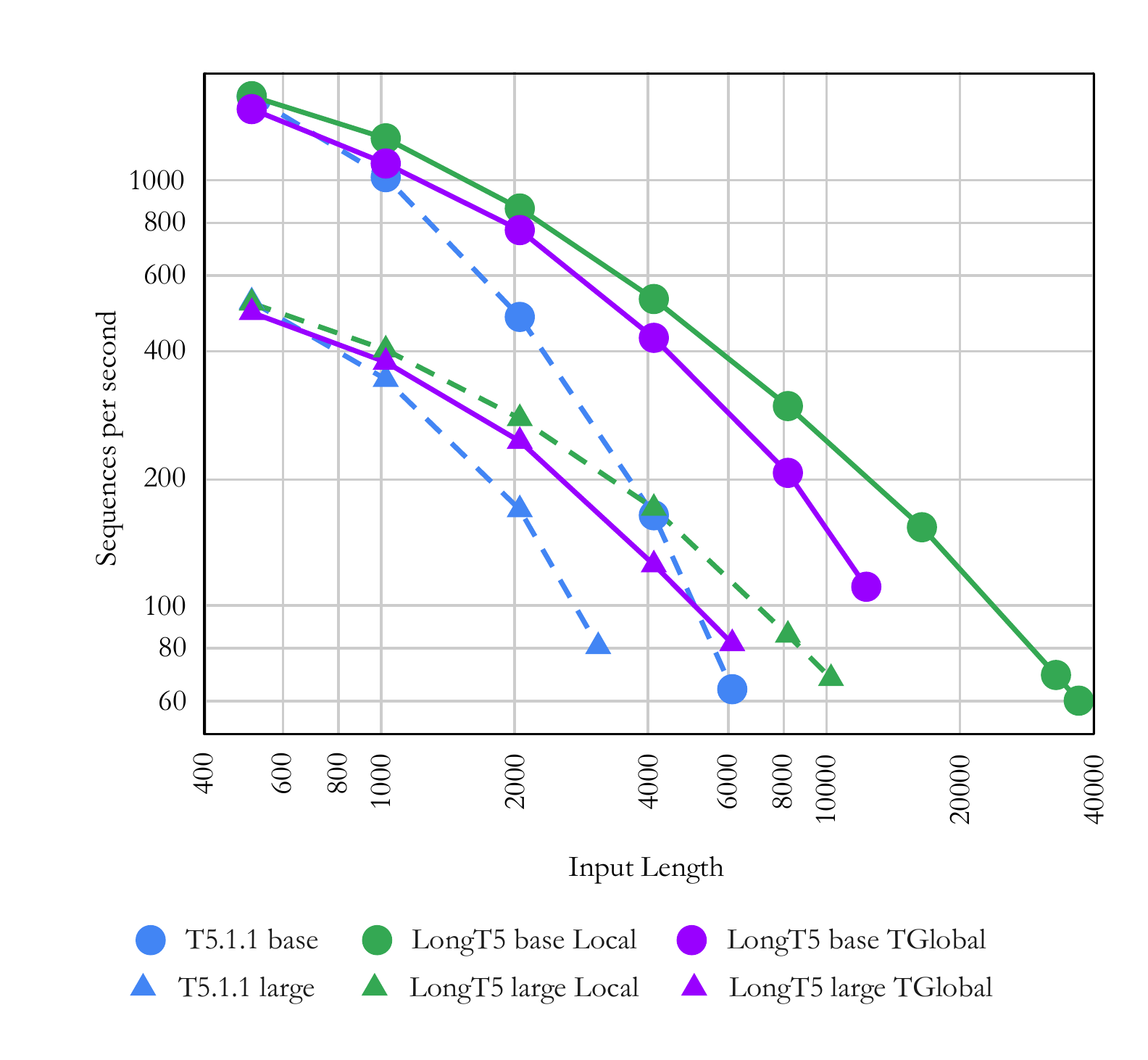}
	\centering
	\caption{Sequences per second as a function of input length for T5.1.1, LongT5 with Local Attention and LongT5 with TGlobal attention. Input lengths start at 512, and go as far as possible before running out of memory. Measurements taken with batch size 128, on 4x8 TPUv3 slices. {\em base} and {\em large} model sizes shown.}
	\label{fig:speed}
\end{figure}

In order to evaluate the training speed and memory consumption of LongT5, compared to T5.1.1, we performed a series of training runs in the NQ data set starting at input length 512, and increasing the input length steadily until models ran out of memory on a 4x8 TPUv3 slice. Results are shown in Figure \ref{fig:speed}, which compares 6 different model configurations: T5.1.1 base, T5.1.1 large, LongT5 (base Local), LongT5 (large Local), LongT5 (base TGlobal), and LongT5 (large TGlobal). For each model configuration, we show a curve plotting the number of sequences per second processed during training (speed, in the vertical axis) for each input length (horizontal axis). Both axes are shown in logarithmic scale.

We can see that at shorter lengths (512), T5.1.1, LongT5 Local, LongT5 TGlobal have similar speeds, but as we increase the sequence length, LongT5 becomes significantly faster. For example at sequence length 2048, T5.1.1 base can only process 479 sequences per second, while LongT5 (base TGlobal) can process 765 and LongT5 (base Local) can process 860. The differences grow even larger as sequence length increases.

Another important fact that Figure \ref{fig:speed} shows is that T5.1.1 models reach their out of memory point much earlier. For example, we could only scale up to 6k tokens for T5.1.1 base. On the other hand, LongT5 (base Local) can go up to 36k tokens in length, and LongT5 (base TGlobal) up to 12k. Large models show a similar picture with T5.1.1 large going only up to 3k, but the LongT5 variants going to 10k (large Local) and 6k (large TGlobal).


\subsection{Input Length vs Performance} \label{subsec:analysis:performance}


\begin{figure}[t!]
	\includegraphics[width=\columnwidth]{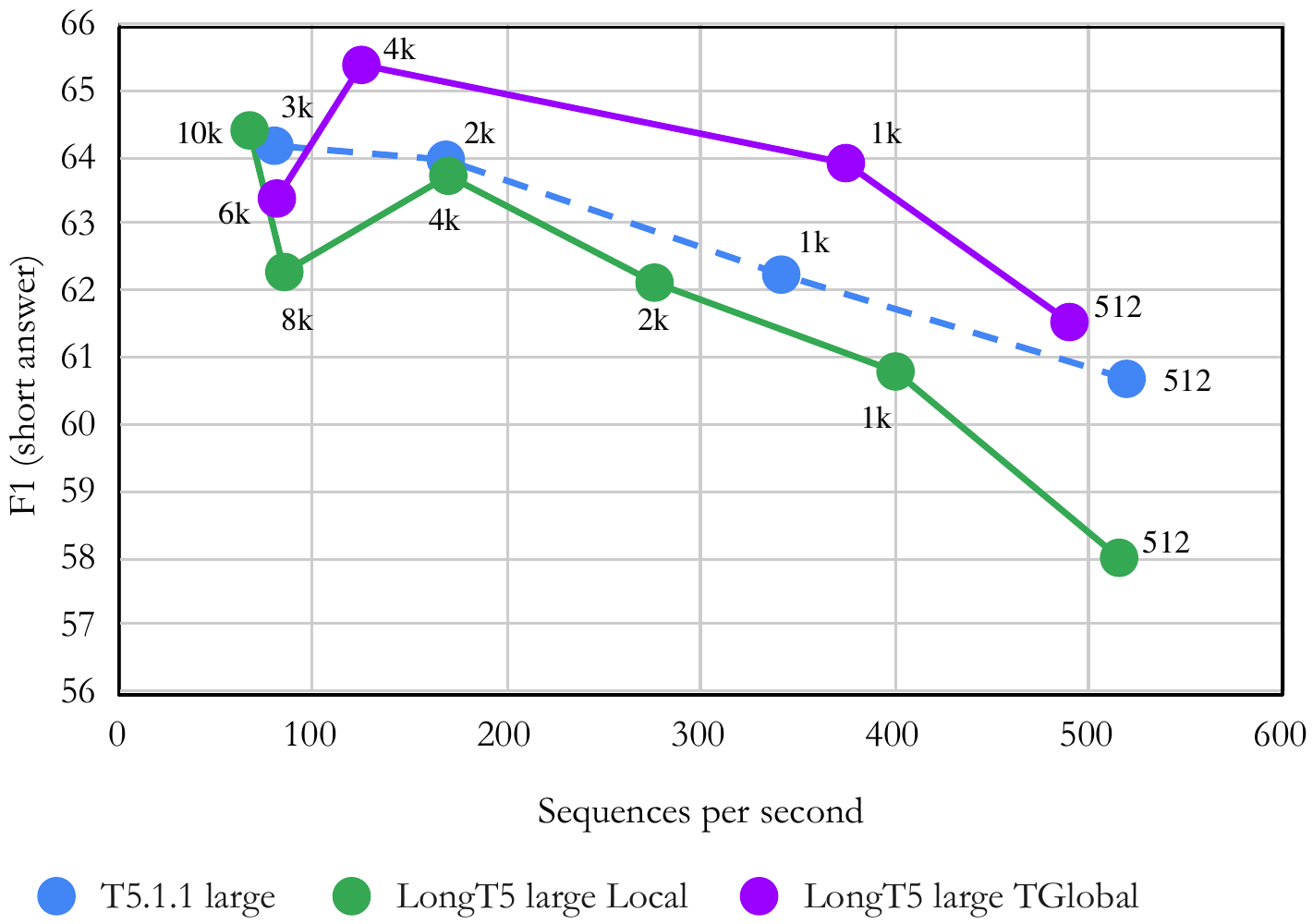}
	\centering
	\caption{Speed versus Performance on NQ (short-answer F1), for  T5, LongT5 with Local Attention and LongT5 with TGlobal attention, for different input sequence lengths. Input lengths start at 512, and go as far as possible before running out of memory. Measurements taken with batch size 128, on 4x8 TPUv3 slices.}
	\label{fig:nq-performance}
\end{figure}

This section presents a similar analysis, but where we plotted model speed versus performance in NQ (F1 score). Results are shown in Figure \ref{fig:nq-performance} for models with {\em large} size. Each point in the curves is annotated with the corresponding sequence length. 

As Figure \ref{fig:nq-performance} shows, performance increases significantly as input length increases, highlighting the benefits of LongT5. Moreover, input length by itself is not enough to achieve good performance in all datasets, and in particular, in the NQ dataset (used in this figure), using Local Attention significantly hurts performance when compared with TGlobal or with T5.1.1. So, even at very long input lengths, LongT5 with Local Attention just matches T5.1.1 with input length of 3k in NQ. However, LongT5 with TGlobal attention outperforms T5.1.1. Moreover, note that although the plot shows a few irregularities (such as 8k length for LongT5 with Local Attention, or 6k length with TGlobal Attention), that is because the plot shows only the results of a single run, and hence there is some noise. However, trends can clearly be seen.


\subsection{Principle Sentences Generation vs. Span Corruption} \label{subsec:analysis:pegasus_vs_span}
As mentioned in section~\ref{sec:pegasus}, we use PEGASUS Principle Sentences Generation instead of default Span Corruption used in T5 as our pre-training objective.
Table~\ref{tab:pre-training} shows our ablation study for fine-tuning on NQ and arXiv from a model pre-trained using the default Span Corruption objective, a model pre-trained with Principle Sentences Generation, and a model pre-trained with both objectives. The comparison is done on the dev set of the tasks, and with TGlobal base models. Both pre-training and fine-tuning on the models mentioned above are done with input sequence length 4096. The table shows, even though Principle Sentences Generation was developed by \citet{pegasus} as a pre-training strategy for summarization, it benefits both summarization and QA tasks, but using both objectives together perform worse than just using PSG.

Table~\ref{tab:pre-training-arxiv} shows an additional ablation study with arXiv and PubMed, where we compare using regular T5.1.1 with Span Corruption compared to T5.1.1 pretrained with Principle Sentences Generation while using the same pre-training input sequence length of 512 (as was done in the original T5.1.1 pre-training task). As expected, Principle Sentences Generation helped the model achieve better results compared to Span Corruption when seeing the same amount of pre-training data.
We also compare this with dev scores from LongT5 with TGlobal attention at 4k and 16k input lengths, such that we can see having full attention will allow for better results, but being able to scale to longer input sequence lengths allows LongT5 to achieve its stronger results.

\begin{table}[t!]
\small
\centering
\begin{tabular}{lccccc}
\toprule
& \multicolumn{2}{c}{\textbf{NQ}} & \multicolumn{3}{c}{\textbf{arXiv}} \\
\textbf{Objective} & EM & F1 & R-1 & R-2 & R-3\\
\midrule
PSG & 62.21 & 66.94 & 44.95 & 18.74 & 40.99\\
\midrule
SC & 58.65 & 63.05 & 43.49 & 18.12 & 39.71 \\
SC + PSG & 59.74 & 64.54 & 44.85 & 18.79 & 40.90 \\
\bottomrule
\end{tabular}
\caption{Ablation study on dev set for different pre-training strategies using span corruption (SC) vs. principle sentences generation (PSG) and the effects on NQ and arXiv fine-tuning tasks. The models are TGlobal base, and fine-tuning is done with input sequence length 4096.}
\label{tab:pre-training}
\end{table}

\begin{table}[t!]
\small
\centering
\begin{tabular}{lccc}
\toprule
& \multicolumn{3}{c}{\textbf{arXiv}} \\
\textbf{Objective} & R-1 & R-2 & R-3\\
\midrule
SC & 44.59 & 18.34 & 40.65 \\
PSG & 45.78 & 18.94 & 41.53 \\
LongT5 (4k) & 45.66 & 19.22 & 41.49 \\
LongT5 (16k) & 48.21 & 21.7 & 44.03 \\
\midrule
& \multicolumn{3}{c}{\textbf{PubMed}} \\
\textbf{Objective} & R-1 & R-2 & R-3\\
\midrule
SC & 47.86 & 22.14 & 44.39 \\
PSG & 48.74 & 23.42 & 45.24 \\
LongT5 (4k) & 48.47 & 23.38 & 45.01 \\
LongT5 (16k) & 50.12 & 24.78 & 46.56 \\
\bottomrule
\end{tabular}
\caption{Ablation study on arXiv and PubMed for different pre-training strategies using span corruption (SC) vs. principle sentences generation (PSG) with T5.1.1 model along with LongT5 with TGlobal attention. Fine-tuning was done on large model size, with input sequence length of 4096 except where otherwise noted.}
\label{tab:pre-training-arxiv}
\end{table}


\section{Related Work}\label{sec:related}

\paragraph{Language model pre-training} 
followed by task specific fine-tuning has proven to be a powerful tool for numerous NLP tasks~\cite{devlin-etal-2019-bert,liu2019roberta,zhang-etal-2019-ernie,Radford2019LanguageMA,t5,lewis-etal-2020-bart,joshi-etal-2020-spanbert}.
BERT~\cite{devlin-etal-2019-bert} introduced Mask Language Model (MLM), where a model predicts masked tokens given a sequence of text input. Fine-tuning a pre-trained BERT model has led to improved performance on various NLP tasks.
However, MLM predictions are not made auto-regressively, which limits the capability of the BERT family for generation tasks.
\citet{t5} introduced the span corruption task in T5 as the pre-training objective, where a model predicts the masked token span using an autoregressive model.
It can handle the generation tasks as the pre-training is done in a generative way.
BART~\cite{lewis-etal-2020-bart} is similar to T5 but used a slightly different pre-training objective, in which spans are masked from the input but the complete output is predicted. 
However, none of these works tried to investigate pre-training for very long sequence inputs.
They often use a transformer~\cite{transformer} architecture as backbone, the complexity of which is quadratic to the input length, making them impractical to model very long sequence input.


\paragraph{Long text modeling}
An extensive amount of work has also been done for modeling long text like documents. The work from~\citet{roy2016,chen2017,wu2018} obtained document embeddings from word-level embeddings. Another line of research tries to model long documents through hierarchical training. The work from \citet{yang-etal-2016-hierarchical,miculicich-etal-2018-document} employed Hierarchical Attention Networks for document classification and neural machine translation, and \citet{guo-etal-2019-hierarchical} proposed using a hierarchy network to build document embeddings on top of sentence embeddings for parallel document mining. 

More recent research has been focusing on improving the memory and computation efficiency of transformer models~\cite{Tay2020EfficientTA,tay2021long} for handling long input. 
One type of such approaches is using non-full attention patterns to restrict the attention field range, so that it reduces the attention complexity from $O(n^2)$ to $O(n log n)$ or $O(n)$, including Sinkhorn~\cite{Tay2020SparseSA}, Longformer~\cite{Beltagy2020Longformer}, ETC~\cite{etc}, and BigBird~\cite{bigbird}.
Another type of approaches is leveraging the low-rank approximation of the attention matrix, such as Linformer~\cite{wang2020linformer}, Performer~\cite{choromanski2021rethinking}, Random Feature Attention~\cite{peng2021random}, and LUNA~\cite{ma2021luna}.


\section{Conclusion}

This paper presented a new Transformer-based neural model called {\em LongT5}, with which we have explored the effects of scaling both input length and model size at the same time. Specifically, the main differences of LongT5 with respect to T5.1.1 are (1) a new scalable attention mechanism called {\em Transient Global} attention, which is a drop-in replacement to the standard T5 attention mechanism, and hence can be used without needing additional side-inputs to the model or modifications to the model inputs; and (2) using a PEGASUS-style Principle Sentences Generation pre-training objective.

Via experimentation in several challenging summarization and question answering datasets, we have explored the performance gains that can be achieved by scaling both input length and model size, resulting in state-of-the-art results on several datasets: arXiv, PubMed, BigPatent, MediaSum, and TriviaQA.

As part of our future work, we would like to pursue several directions such as studying efficient attention mechanisms in the decoder and decoder-to-encoder attention pieces of the model (both Local Attention and TGlobal attention are only applied to the encoder in LongT5 for now). Additionally, we would like to incorporate additional long-input transformer ideas into the LongT5 architecture, that could further improve model efficiency.



\clearpage

\appendix

\section{Summarization Results}\label{sec:sum_appendix}

Table \ref{tab:summarization_full_results} shows the full set of results on the summarization datasets used in this paper.
This includes both standard T5 model (using version T5.1.1), T5 with PEGASUS Principle Sentences Generation pre-training, and LongT5 model.

As can be seen, scaling up the input size for the models helps achieve better performance metrics.
T5 models though struggle when scaling up to 4k for input, as the fine-tuning task can take many days even when using a large topology of TPUv3.

When comparing regular T5.1.1 model with a T5.1.1 model using PEGASUS Principle Sentences Generation pre-training, the latter was able to achieve better results, with the results also improving as the input size scaled up.
This helps show that both using the latter pre-training objective along with scaling up allows us to get the best results from these models.

LongT5, despite having a reduced attention from using TGlobal attention, is able to get strong performance results due to both scaling up to larger inputs and leveraging the Gap Sentences Generation pre-training strategy.

\begin{table}
\small
\centering
\begin{tabular}{lcccc}
\toprule
& \multicolumn{2}{c}{\textbf{NQ}} & \multicolumn{2}{c}{\textbf{TriviaQA}} \\
\textbf{Approach} & EM & F1 & EM & F1\\
\midrule
{\bf base:} \\
T5.1.1 (512) & 50.93 & 52.54 & 48.91 & 52.89 \\
T5.1.1 (6k) & 56.73 & 56.73 & 59.09 & 63.31 \\
{\bf large:} \\
T5.1.1 (512) & 57.29 & 60.68 & 53.26 & 57.01 \\
T5.1.1 (3k) & 60.09 & 64.17 & 60.15 & 64.15 \\
{\bf xl:} \\
T5.1.1 (4k) & 60.75 & 64.07 & 65.33 & 69.43 \\
\midrule
{\bf base Local:} \\
LongT5 (512) & 54.39 & 58.24 & - & - \\
LongT5 (1k) & 54.60 & 57.88 & - & - \\
LongT5 (2k) & 56.48 & 60.56 & - & - \\
LongT5 (4k) & 56.10 & 60.52 & - & - \\
LongT5 (8k) & 55.90 & 59.98 & - & - \\
LongT5 (16k) & 56.41 & 60.46 & - & - \\
LongT5 (32k) & 55.84 & 59.59 & - & - \\
LongT5 (36k) & 55.77 & 59.66 & - & - \\
{\bf base TGlobal:} \\
LongT5 (512) & 55.73 & 59.06 & - & - \\
LongT5 (1k) & 57.41 & 61.25 & - & - \\
LongT5 (2k) & 56.96 & 60.25 & - & - \\
LongT5 (4k) & 58.97 & 63.03 & - & - \\
LongT5 (8k) & 58.07 & 62.67 & - & - \\
LongT5 (12k) & 58.12 & 62.44 & 63.27 & 67.42 \\
{\bf large Local:} \\
LongT5 (512) & 55.19 & 58.00 & - & - \\
LongT5 (1k) & 57.47 & 60.79 & - & - \\
LongT5 (2k) & 58.49 & 62.12 & - & - \\
LongT5 (4k) & 59.44 & 63.72 & - & - \\
LongT5 (8k) & 58.66 & 62.28 & - & - \\
LongT5 (10k) & 60.01 & 64.40 & - & - \\
{\bf large TGlobal:} \\
LongT5 (512) & 57.55 & 61.53 & - & - \\
LongT5 (1k) & 59.69 & 63.91 & - & - \\
LongT5 (4k) & 60.77 & 65.38 & - & - \\
LongT5 (6k) & 59.17 & 63.38 & 63.76 & 67.82 \\
{\bf xl TGlobal:} \\
LongT5 (4k) & 62.38 & 66.39 & - & -\\
LongT5 (8k) & {\bf 62.66} & {\bf 66.61} & {\bf 67.89} & {\bf 71.71} \\
\bottomrule
\end{tabular}
\caption{QA results comparing T5.1.1 and LongT5 at different sequence lengths. Base and large models are trained on 4x8 TPUv3 with no model partitioning, and xl models are trained on 8x16 TPUv3 with 8 partitions.}
\label{tab:QA_results_full}
\end{table}

\begin{table*}
\small
\centering
\begin{tabular}{lcccccc}
\toprule
& \multicolumn{3}{c}{\textbf{arXiv}} & \multicolumn{3}{c}{\textbf{PubMed}} \\
\textbf{Approach} & R-1 & R-2 & R-L & R-1 & R-2 & R-L\\
\midrule
DANCER PEGASUS & 45.01 & 17.6 & 40.56 & 46.34 & 19.97 & 42.42 \\
BigBird-PEGASUS (large) & 46.63 & 19.02 & 41.77 & 46.32 & 20.65 & 42.33  \\
HAT-BART & 46.68 & 19.07 & 42.17 & 48.36 & 21.43 & 37.00 \\
LED (large) & 46.63 & 19.62 & 41.83 & - & - & -  \\ 
PRIMER & 47.6 & 20.8 & 42.6 & - & - & - \\
\midrule
T5.1.1 (large - 1k input) & 39.79 & 14.02 & 36.23 & 42.18 & 16.60 & 38.96 \\
T5.1.1 (large - 2k input) & 42.84 & 16.62 & 39.01 & 45.51 & 19.55 & 42.10 \\
T5.1.1 (large - 4k input) & 44.51 & 18.20 & 40.62 & 47.90 & 22.08 & 44.36 \\
T5.1.1 + PSG (large - 1k input) & 38.53 & 13.61 & 35.08 & 43.34 & 17.55 & 40.10 \\
T5.1.1 + PSG (large - 2k input) & 42.85 & 16.50 & 38.99 & 46.51 & 20.37 & 43.00 \\
T5.1.1 + PSG (large - 4k input) & 45.86 & 18.40 & 41.62 & 48.94 & 22.92 & 45.4 \\
LongT5 (base - 4k input) & 44.87 & 18.54 & 40.97 & 47.77 & 22.58 & 44.38 \\
LongT5 (large - 4k input) & 45.64 & 18.6 & 41.51 & 48.38 & 23.32 & 44.93 \\
LongT5 (large - 8k input) & 46.61 & 19.67 & 42.44 & 49.81 & 24.3 & 46.26 \\
LongT5 (large - 16k input) & 48.28 & 21.63 & 44.11 & 49.98 & 24.69 & 46.46 \\
LongT5 (xl - 4k input) & 45.99 & 19.51 & 42.04 & 48.99 & 23.48 & 45.51 \\
LongT5 (xl - 8k input) & 47.44 & 20.84 & 43.34 & 50.04 & 24.45 & 46.42 \\
LongT5 (xl - 16k input) & \textbf{48.35} & \textbf{21.92} & \textbf{44.27} & \textbf{50.23} & \textbf{24.76} & \textbf{46.67} \\
\toprule
& \multicolumn{3}{c}{\textbf{BigPatent}} & \multicolumn{3}{c}{\textbf{MultiNews}} \\
\textbf{Approach} & R-1 & R-2 & R-L & R-1 & R-2 & R-L\\
\midrule
BigBird-PEGASUS (large) & 60.64 & 42.46 & 50.01 & - & - & -  \\
TG-MultiSum & - & - & - & 47.10 & 17.55 & 20.73 \\
PRIMER & - & - & - & \textbf{49.9} & \textbf{21.1} & \textbf{25.9} \\
\midrule
T5.1.1 (large - 1k input) & 55.07 & 37.49 & 45.90 & 43.69 & 16.26 & 23.03 \\
T5.1.1 (large - 2k input) & 60.07 & 43.49 & 50.90 & 44.95 & 17.26 & 23.74 \\
T5.1.1 (large - 4k input) & 62.14 & 45.85 & 52.95 & 45.67 & 17.88 & 24.15 \\
T5.1.1 + PSG (large - 1k input) & 58.58 & 41.80 & 49.74 & 44.43 & 15.85 & 22.41 \\
T5.1.1 + PSG (large - 2k input) & 64.51 & 49.15 & 56.01 & 46.65 & 17.74 & 23.74 \\
T5.1.1 + PSG (large - 4k input) & 67.05 & 52.24 & 58.70 & 47.48 & 18.60 & 24.31 \\
LongT5 (base - 4k input) & 60.95 & 44.22 & 51.52 & 46.01 & 17.37 & 23.5 \\
LongT5 (large - 4k input) & 66.17 & 51.10 & 57.70 & 46.99 & 18.21 & 24.08 \\
LongT5 (large - 8k input) & 67.42 & 52.62 & 59.04 & 47.18 & 18.44 & 24.18 \\
LongT5 (large - 16k input) & 70.38 & 56.81 & 62.73 & - & - & - \\
LongT5 (xl - 4k input) &  75.82 & 64.64 & 69.54 & 48.15 & 19.30 & 24.76 \\
LongT5 (xl - 8k input) & 76.39 & 65.37 & 70.16 & 48.17 & 19.43 & 24.94 \\
LongT5 (xl - 16k input) & \textbf{76.87} & \textbf{66.06} & \textbf{70.76} & - & - & - \\
\toprule
& \multicolumn{3}{c}{\textbf{MediaSum}} & \multicolumn{3}{c}{\textbf{CNN / Daily Mail}} \\
\textbf{Approach} & R-1 & R-2 & R-L & R-1 & R-2 & R-L\\
\midrule
HAT-BART & - & - & - & \textbf{44.48} & 21.31 & \textbf{41.52} \\
BART (large) & 35.09 & 18.05 & 31.44 & - & - & -   \\
\midrule
T5.1.1 (large - 1k input) & 30.68 & 14.88 & 27.88 & 42.60 & 20.41 & 40.03 \\
T5.1.1 (large - 2k input) & 32.83 & 16.75 & 29.79 & 42.55 & 20.25 & 39.99 \\
T5.1.1 (large - 4k input) & 34.37 & 18.09 & 31.12 & 42.27 & 19.93 & 39.72 \\
T5.1.1 + PSG (large - 1k input) & 32.02 & 16.15 & 28.89 & 42.62 & 20.46 & 40.02 \\
T5.1.1 + PSG (large - 2k input) & 34.04 & 17.87 & 30.77 & 42.69 & 20.40 & 40.06 \\
T5.1.1 + PSG (large - 4k input) & 36.11 & 19.48 & 32.67 & 43.41 & 20.99 & 40.77 \\
LongT5 (base - 4k input) & 35.09 & 18.35 & 31.87 & 42.15 & 20.11 & 39.6 \\
LongT5 (large - 4k input) & 35.54 & 19.04 & 32.20 & 42.49 & 20.51 & 40.18 \\
LongT5 (xl - 4k input) & \textbf{36.15} & \textbf{19.66} & \textbf{32.80} & 43.94 & \textbf{21.40} & 41.28 \\
\bottomrule
\end{tabular}
\caption{Summarization results comparing T5, T5 with PEGASUS-style Principle Sentences Generation (PSG) pre-training, and LongT5 with best known approaches for the various datasets. All T5 scores are with standard T5.1.1 model. All LongT5 scores are with models using TGlobal attention. For each task, we scale up the input length depending on the statistics of the inputs, thus not all of the tasks were scaled to 16k. We do not include input length of other models because each model uses the input differently, and hence, direct comparison is not possible.}
\label{tab:summarization_full_results}
\end{table*}

\section{QA Results}\label{sec:qa_appendix}

Table \ref{tab:QA_results_full} shows the full set of results comparing T5.1.1 and LongT5 models on the QA datasets used in this paper. For both NQ and TriviaQA in this comparison study, we use 90\% of the official training set for training while using 10\% as hold-out dev set to fine-tune the hyperparameters and training epoch, and use the official dev set to report the numbers in this table. We run each model to the largest input length allowed before running out of memory on specific hardware configuration - base/large models on 4x8 TPUv3 with no model partitioning, and xl models on 8x16 TPUv3 with 8 partitions.


\end{document}